\documentclass{ieeeaccess}
\usepackage{cite}
\usepackage{amsmath,amssymb,amsfonts}
\usepackage{algorithm}
\usepackage{algpseudocode}
\usepackage{graphicx}
\usepackage{textcomp}
\usepackage{booktabs}
\usepackage{array}

\usepackage{microtype}
\usepackage{graphicx}
\usepackage{booktabs} 
\usepackage{subfig}

\usepackage{xspace}
\usepackage{amsmath}
\usepackage{amssymb}
\usepackage{amsfonts}
\usepackage{bm}
\usepackage{bbm}
\usepackage{float}
\usepackage{stfloats}
\usepackage{wrapfig}
\usepackage[pagebackref=true,breaklinks=true,colorlinks,bookmarks=false]{hyperref}
\usepackage{multirow}
\usepackage{lipsum}
\usepackage{mathtools}
\usepackage{cuted}
\usepackage{stfloats}

\definecolor{Green}{RGB}{102,252,102}

\newcommand{\ie}{\textit{i.e. }}

\usepackage{mathtools,amssymb,lipsum}
\newcommand{\PreserveBackslash}[1]{\let\temp=\\#1\let\\=\temp}
\newcolumntype{C}[1]{>{\centering\arraybackslash}p{#1}}

\def\BibTeX{{\rm B\kern-.05em{\sc i\kern-.025em b}\kern-.08em
    T\kern-.1667em\lower.7ex\hbox{E}\kern-.125emX}}

\begin{document}
\history{Date of publication 11 January 2023, date of current version 11 January 2023.}
\doi{10.1109/ACCESS.2023.3236087}

\title{DimCL: Dimensional Contrastive Learning For Improving Self-Supervised Learning}

\author{\uppercase{Thanh Nguyen }\authorrefmark{$\mathbf{\dagger}$1}, \uppercase{Trung Pham}\authorrefmark{$\mathbf{\dagger}$1},  \uppercase{Chaoning Zhang}\authorrefmark{1},  \uppercase{Tung Luu}\authorrefmark{1}, \uppercase{Thang Vu}\authorrefmark{1} and \uppercase{Chang D. Yoo \authorrefmark{1}}}

\address[1]{School of Electrical Engineering, Korea Advanced Institute of Science and Technology, Daejeon 34141, Republic of Korea}

\tfootnote{$^{\dagger}$ Equal contribution.\\This work was supported by the Institute for Information \& communications Technology Promotion (IITP) grant funded by the Korea government (MSIT) (No. 2021-0-01381,
Development of Causal AI through Video Understanding and Reinforcement Learning, and Its Applications to Real Environments and No.2022-0-00184, Development and Study of AI Technologies to Inexpensively Conform to Evolving Policy on Ethics).}

\markboth
{Author \headeretal: Preparation of Papers for IEEE TRANSACTIONS and JOURNALS}
{Author \headeretal: Preparation of Papers for IEEE TRANSACTIONS and JOURNALS}

\corresp{Corresponding author: Chang D. Yoo (e-mail: cd\_yoo@kaist.ac.kr).}

\begin{abstract}
Self-supervised learning (SSL) has gained remarkable success, for which contrastive learning (CL) plays a key role. However, the recent development of new non-CL frameworks has achieved comparable or better performance with high improvement potential, prompting researchers to enhance these frameworks further. Assimilating CL into non-CL frameworks has been thought to be beneficial, but empirical evidence indicates no visible improvements. In view of that, this paper proposes a strategy of performing CL along the dimensional direction instead of along the batch direction as done in conventional contrastive learning, named Dimensional Contrastive Learning (DimCL). DimCL aims to enhance the feature diversity, and it can serve as a regularizer to prior SSL frameworks. DimCL has been found to be effective, and the hardness-aware property is identified as a critical reason for its success. Extensive experimental results reveal that assimilating DimCL into SSL frameworks leads to performance improvement by a non-trivial margin on various datasets and backbone architectures.

\end{abstract}

\begin{keywords}
Self-supervise learning, Computer Vision, Contrastive Learning, Deep Learning, Transfer Learning.
\end{keywords}
\titlepgskip=-15pt
\maketitle
\section{Introduction}

The success of self-supervised learning (SSL) has been demonstrated in a wide range of applications, ranging from early attempts in natural language processing \cite{Lan2020ALBERT,radford2019language,devlinetal2019bert,su2020vlbert,nie2020dc} to more recent computer vision tasks \cite{li2021esvit,chen2021mocov3,el2021xcit}. To be more specific, in contrast to supervised learning which requires a huge amount of labeled data \cite{5206848,zhai2022scaling,vu2022softgroup,vu2022softgroup}, SSL learns the representations without the need for labeled ones. Thus, it significantly reduces the human-label cost and enables machine learning to learn from a massive amount of available unlabeled data leading to benefits for many real-world applications in various fields: teaching robots to work from raw pixel images \cite{laskin2020curl,nguyen2021sample,luu2022visual,luu2022utilizing}, training medical diagnosis systems from un-labels checkups results \cite{krishnan2022self,chen2019self}, enhance 3D face reconstruction using images in the wild \cite{tu20203d}, ...

Without using human annotation labels, SSL methods seek to learn an encoder  with augmentation-invariant representation  \cite{bachman2019learning,he2020momentum,chen2020simple,caron2020unsupervised,grill2020bootstrap}. A common approach is to minimize the distance between two representations of positive samples, \ie two augmented views of the same image. Based on this simple approach, the past few years have witnessed the development of various SSL frameworks, which can be roughly divided into two categories: CL-based and non-CL frameworks. The CL-based frameworks \cite{oord2018representation,hjelm2018learning,wu2018unsupervised,zhuang2019local,bachman2019learning,henaff2020data,tian2020contrastive,chen2020simple,he2020momentum,wang2020understanding,wang2020DenseCL} have achieved remarkable developments and greatly contributed to the progress of  SSL. Recently, multiple works \cite{chen2021exploring,grill2020bootstrap,ermolov2021whitening,zbontar2021barlow,bardes2021vicreg} have also demonstrated successful attempts with the non-CL frameworks, among which BYOL \cite{chen2021exploring} and SimSiam \cite{chen2020simple} are the two representatives.

\begin{figure*}[!tp] 
    \centering
    \includegraphics[width=0.8\textwidth]{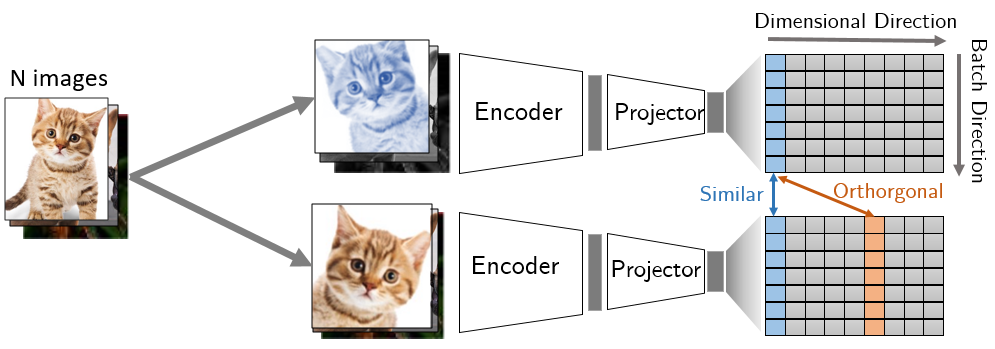}
         \caption{Dimensional contrastive learning (DimCL). 
         As the term suggests, 
         existing BCL performs CL along the batch direction to encourage diversity of representations, while our proposed DimCL performs CL along the dimensional direction to encourage diversity among elements within a representation (termed feature diversity). Our DimCL can be used as a plug-and-play regularization method to improve non-CL (and CL-based) SSL frameworks.
    }
  \label{fig:general_arch}
\end{figure*}

Compared with the CL-based frameworks, the non-CL ones~\cite{chen2021exploring,chen2020simple} have a unique advantage: they propose simpler frameworks without using the negative samples, yet achieve comparable or even superior performance on benchmark datasets (like ImageNet-1K and CIFAR-10/100). Thus, there is a trend to shift from CL to non-CL frameworks. Recognizing the significance of CL in the development of SSL, this work attempts to distill beneficial properties of CL to push the frontiers of non-CL frameworks further. However, naively assimilating CL to non-CL does not show visible improvement, as pointed out in BYOL \cite{grill2020bootstrap}. 
This can be attributed to the fact that the frameworks mentioned above focus on the same inter-instance level of constraints and mainly pursue the same objective (augmentation invariant). In essence, existing CL encourages representation diversity among the instances in the batch. In this paper, CL is utilized to encourage diversity among the representation elements in obtaining ``\textit{feature diversity}'', referred to as Dimensional Contrastive Learning.
To avoid any confusion between batch contrastive learning and dimensional contrastive learning, we denote each as BCL and DimCL, respectively. The difference between BCL and DimCL is depicted in Fig. \ref{fig:compare_bcl_dcl}.

A prudent variation in BCL led to a separate SSL framework, while the proposed DimCL (as illustrated in Fig. \ref{fig:general_arch}) is designed as a regularizer for feature diversity enhancement to support other frameworks. Even though DimCL is originally motivated to boost non-CL frameworks, empirically, DimCL is found to  also enhance the performance of existing CL-based frameworks and can be generalized to other domains (e.g,. supervised learning). This implies that feature diversity is necessary for good representations. 

Our contributions are as follows:
\begin{itemize}
    \item Recognizing the significance of CL in the development of self-supervised learning, we are the first to apply DimCL to push the frontiers of non-CL frameworks. In contrast to existing BCL, our proposed DimCL performs CL along the dimensional direction and can be used as a regularizer for boosting the performance of non-CL (and CL-based) frameworks. 
    \item We perform extensive experiments on various frameworks with different backbone architectures on diverse datasets to validate the effectiveness of our proposed DimCL. We also investigate the reason for the benefit brought by DimCL and identify the hardness-aware property as an essential factor.  
\end{itemize}

The rest of this paper is organized as follows. Section II summarizes the related works. Section III describes the background of Batch Contrastive Learning. Section IV presents the proposed method DimCL. Section V provides the experiment setup and results. Section VI shows the ablation study on important hyper-parameters. Section VII provides some discussions about DimCL. Finally, Section VII concludes this work.

\section{Related Work}

\textbf{ Contrastive Learning.} Contrastive learning (CL) is one of the prominent keystones of self-supervised learning. It fosters discriminability in the representation \cite{Schroff2015FaceNetAU, wang2015unsupervised, sohn2016improved, misra2016shuffle, Federici2020Learning}. Early works have studied margin-based contrastive losses \cite{hadsell2006dimensionality, wang2015unsupervised, hermans2017defense}. After the advent of \cite{wu2018unsupervised, oord2018representation}, NCE-based loss has become the standard loss in CL. Inspired by this success, CL has been extensively studied for SSL pretext training \cite{wu2018unsupervised, oord2018representation, bachman2019learning,henaff2020data, hjelm2018learning, tian2019contrastive,zhuang2019local,chen2020simple}. SimCLR \cite{chen2020simple} proposes a simple yet effective method to train the unsupervised model. They show that more negative samples (4096, for instance) are beneficial for performance. However, such a massive number of negative samples require a huge batch size for training to achieve the desired performance.

MoCo v1 \cite{he2020momentum} has attracted significant attention by demonstrating superior performance over supervised pre-training counterparts in downstream tasks while making use of large negative samples, decoupling the need for batch size by introducing a dynamic dictionary. Inspired by \cite{chen2020simple}, MoCo v2 \cite{chen2020mocov2} applies stronger augmentations and an additional MLP projector, which shows significant performance improvement over the first version of MoCo. \cite{chen2021empirical} has empirically shown that the predictor from the non-CL frameworks \cite{chen2021exploring,grill2020bootstrap} helps to gain performance boost for MoCo variants with ViT structures \cite{dosovitskiy2021an}.

Several works explain the key properties that lead to the success of CL. It is noticeable that momentum update \cite{chen2020simple} and large negative samples play an important role in preventing collapse. InfoNCE loss was identified to have the hardness-aware property, which is critical for optimization \cite{Wang_2021_CVPR} and preventing collapse by instance de-correlation \cite{anonymous2022how}. \cite{nozawa2021understanding, Vasudeva_2021_ICCV, Wang_2021_ICCV, iscen2018mining, chuang2020debiased, ho2020contrastive, wu2020conditional} have demonstrated that hard negative samples mining strategies can be beneficial for better performance over the baselines. Notably, \cite{wang2020understanding} identified CL form alignment and uniformity of feature space which benefits downstream tasks.

Most of the contrastive learning frameworks adopt the instance discrimination task which inevitably causes class collision problems \cite{zheng2021weakly} where the representations of the same class images are forced to be different.  The problem can hurt the quality of the learned representation. Different from the above methods, which perform CL along the batch direction, DimCL performs the CL along the dimensional direction in order to encourage diversity among representation elements instead of representation vectors. This approach never faces class collision problems.

\textbf{ Non-Contrastive Learning.}  Non-contrastive learning focuses on making augmentation invariant without using negative samples. With the absence of negative samples, training the simple siamese network using the cosine similarity loss leads to complete collapse \cite{anonymous2022how, grill2020bootstrap}. BYOL \cite{grill2020bootstrap} and Simsiam \cite{chen2021exploring} demonstrated that using a careful architecture design to break the architecture symmetry can avoid collapse. Specifically, a special `predictor' network is added in conjunction with the exponential moving average update (BYOL) or with a stop gradient in one branch (Simsiam). Besides, several works have attempted to demystify the success of BYOL \cite{grill2020bootstrap}. A recent work \cite{byol-bn-blog} has suggested that batch normalization (BN) plays a critical role in the success of BYOL; however, another work \cite{richemond2020byol} refutes that claim by showing BYOL works without the need for BN.

Recognizing the strong points of CL in the development process, this work tries to distill beneficial properties of CL in a novel manner and use it as a regularizer to boost the performance of non-CL (and CL) based frameworks. Moreover, most non-CL frameworks aim to learn augmentation invariant representation which training often leads to trivial constant solutions (i.e., collapse) \cite{barlow2001redundancy}. DimCL naturally avoids collapse as it encourages diversity in the solution which is a great complement to non-CL. 

\section{Background}


\begin{figure}\
    \centering
    \centering
    \footnotesize
    \includegraphics[width=0.9\linewidth]{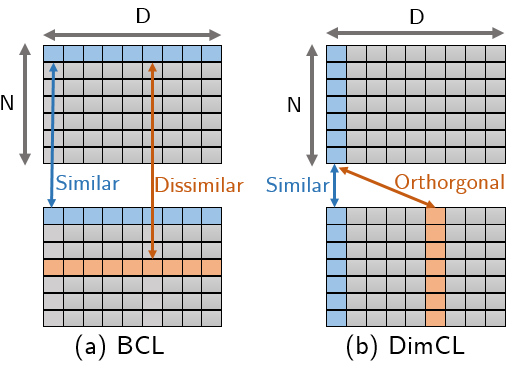}
    \label{fig:BCL_implement)}
    \caption{  The difference between (a) Batch Contrastive Learning (BCL) and (b) Dimensional Contrastive Learning (DimCL) 
    . BCL performs along the batch direction to encourage representation diversity whereas DimCL performs along the dimensional direction to encourage feature diversity. N is the batch size, and D is the feature dimension.}
    \label{fig:compare_bcl_dcl}
\end{figure}

Conventional contrastive learning, \ie BCL, aims to make the representations similar if they are from different augmented versions of the same image and dissimilar if they are from different images. Or shortly, it aims to make meaningful discriminative representations. To be more specific, in BCL, there are query, positive, and negative samples. The considered image is called the query sample. The augmentation views of the query image are called positive samples.  The other images in the sample batch and their augmentation views are called negative samples. The loss of CL-based frameworks basically makes the query representation to be near the positive sample presentation and far apart from the negative sample representation.  Mathematically, given an encoder $f$, an input image is augmented and  encoded as a query $q \in \mathbb{R}^D$ or positive key $k^+ \in \mathbb{R}^D$, which are often $l_2$-normalized to avoid scale ambiguity \cite{he2019moco, chen2020simple}. Processing a mini-batch of N images will form a set of queries  $\mathbb{Q} = \{q_1, q_2, ..., q_N\}$ and positive keys $\mathbb{K}^+ = \{k_1^+, k_2^+, ..., k_N^+\}$. Consider a query $q_i$, the corresponding negative keys are defined $\mathbb{K}^-_i = \mathbb{Q} \cup \mathbb{K}^+ \setminus \{q_i,k^+_i\} =\{k_1^-, k_2^-, ..., k_{2N-2}^-\}$  \cite{chen2020simple}. With similarity measured by dot product, BCL can be achieved by the simple CL loss below \cite{Wang_2021_CVPR}:

\begin{equation}
\begin{aligned}
    \mathcal{L} &= \frac{1}{N} \sum_{i=1}^N \mathcal{L}_i \\
    \mathcal{L}_i &= -q_i{\cdot}k_i^+ + \frac{1}{2N-2}\sum_{j=1}^{2N-2} q_i{\cdot}k_j^-.
    \end{aligned}
        \label{eq:simple} 
\end{equation}
The gradient of $\mathcal{L}_i$ w.r.t $q_i$ is derived as:
\begin{equation}
    \frac{ \partial \mathcal{L}_i }{\partial q_i}  = -k_i^+ + \frac{1}{2N-2}\sum_{j=1}^{2N-2} k_j^-.
    \label{eq:simple_grad}
\end{equation}

The above equation treats all negative keys equally. Based on this, \cite{Wang_2021_CVPR} proved that the superficial loss in Eq. \ref{eq:simple} performs poorly in practice.

The NCE-based loss \cite{gutmann2010noise, oord2018representation} has been independently developed with various motivations in multiple popular works \cite{sohn2016improved, wu2018unsupervised} and it has become the standard loss for BCL. Following \cite{oord2018representation, he2020momentum, anonymous2022how}, we term it InfoNCE for consistency. 
The InfoNCE is formulated as follows \cite{he2019moco}:
\begin{equation}
\begin{aligned}
    \mathcal{L}^{BCL} &= \frac{1}{N} \sum_{i=1}^N \mathcal{L}^{BCL}_i \\
    \mathcal{L}_{{i}}^{BCL} &= -\log \frac{\exp(q_i{\cdot}k_i^+/\tau)}{\exp(q_i{\cdot}k_i^+/\tau) + \sum_{j=1}^{2N-2} \exp(q_i{\cdot}k_j^-/\tau)},
\end{aligned}
        \label{eq:infonce}
\end{equation}
with $\tau$ denoting the temperature.
The InfoNCE has been identified to outperform the above simple loss Eq. \ref{eq:simple} due to its hardness-aware property, which puts more weight on optimizing hard negative pairs (where the query is close to negative keys) as shown in \cite{Wang_2021_CVPR}.

\section{Methodology}

Dimensional Contrastive Learning (DimCL) explores a new way of using InfoNCE compared to BCL. As shown in Fig. \ref{fig:compare_bcl_dcl}, BCL aims to make meaningful discriminative representations by applying InfoNCE along the batch direction. The keys and queries are the representation vectors. By contrast, DimCL encourages each representation element to contain a piece of distinct information to maximize the amount of information contained in the overall representation, which is \textit{feature diversity} enhancement \footnote{Note that feature diversity is defined as the independence among the elements of a representation. It should not be related to diversity among representation vectors}. To this end, the DimCL make the elements of the representation vector orthogonal to each other in term of information by minimizing the empirical correlation among column vectors.  A novel form of InfoNCE  along the dimensional direction is proposed as the loss to achieve this objective. Therein, the corresponding queries and keys are \textit{column vectors}, each of which is formed from the same-index representation elements within a batch, as highlighted in Fig. \ref{fig:compare_bcl_dcl}.

Mathematically, similar to BCL, given a mini-batch of N images, we have a set of queries  $\mathbb{G} = \{g_1, g_2, ..., g_D\}$ and positive keys $\mathbb{H}^+ = \{h_1^+, h_2^+, ..., h_D^+\}$. 
Note that $g,h \in \mathbb{R}^N$ are \textit{column vectors}. Considering a query $g_i$, the corresponding negative keys are defined as $\mathbb{H}^-_i = \mathbb{G} \cup \mathbb{H}^+ \setminus \{g_i,h^+_i\} = \{h_1^-, h_2^-, ..., h_{2D-2}^-\}$.  In order to maximize the feature diversity, the considered query $g_i$ should be orthogonal with all negative keys  $\mathbb{H}^-_i$. The corresponding objective is: 
\begin{equation}
\begin{aligned}
     \mathcal{L}^{AbsCL} &= \frac{1}{D} \sum_{i=1}^D \mathcal{L}_i^{AbsCL} \\
    \mathcal{L}_{{i}}^{AbsCL} &= -\log \frac{\exp(g_i{\cdot} h_i^+ /\tau  )}{\exp(g_i{\cdot}h_i^+/\tau) + \sum_{j=1}^{2D-2} \exp(|g_i {\cdot}h_j^-| /\tau)}.
\end{aligned}
\end{equation}
Empirically, we observe that the original InfoNCE is sufficient to achieve the objective without any modification (e.g., adding the absolute) (evidence is provided in the discussion). This can be explained by considering the $\exp$ term and the effect of temperature  $\tau$. With small $\tau$, the $\exp(x/\tau)$ has high weight on pushing positive value x toward zero with a corresponding high gradient but has almost no consideration on negative value x with the same magnitude due to its much smaller gradient. For simplicity, we adopt the following loss as the DimCL optimization target:
\begin{equation}
\begin{aligned}
         \mathcal{L}^{DimCL} &= \frac{1}{D} \sum_{i=1}^D \mathcal{L}_i^{DimCL} \\
    \mathcal{L}_{{i}}^{DimCL} &= -\log \frac{\exp(g_i{\cdot} h_i^+ /\tau  )}{\exp(g_i{\cdot}h_i^+/\tau) + \sum_{j=1}^{2D-2} \exp(g_i {\cdot}h_j^- /\tau)}.
\end{aligned}
    \label{eq:infodcl}
\end{equation}


Note that, in DimCL each query $g_i$ has total $2D-2$ negative keys instead of $2N-2$ as in BCL. And each of \textit{collumn vector} $g, h$ are $l_2$-normalized along the batch direction instead of dimensional direction as in BCL. 
Furthermore, the proposed DimCL inherits the hardness-aware property of the traditional BCL for which we provide more detail in the discussion part. 

Contrary to BCL, which works as an independent SSL framework, DimCL serves as a regularizer to benefit existing SSL frameworks. We denote $\mathcal{L}^{BASE}$ as the loss of the SSL baseline. DimCL can be simply assimilated into the baseline by a linear combination to form a final loss as:
\begin{equation}
    \mathcal{L} = \lambda  \mathcal{L}^{DimCL} + (1-\lambda)\mathcal{L}^{BASE},
    \label{eq:finalloss}
\end{equation}
where $\lambda \in [0,1]$ is a weight factor to balance the two loss components. We perform a grid search and find 
that $\lambda = 0.1$  works well in most cases and recommend this value as a starting point for more fine-grained tuning.  The pseudo algorithm is provided in Algorithm. \ref{alg:dimCL}

\begin{algorithm}[htb]
    \caption{Pytorch-style pseudocode for DimCL}
    \label{alg:dimCL}
    \begin{algorithmic}
      \State $f$: encoder network, $f'$: target encoder  
      \State $N$: batch size, $D$: dimension 
      \State $\tau$: temperature, $\lambda$: balance weight
      \State $L_{base}$: baseline loss, $Optim$: optimizer 
      
      \For{$x$ {\bfseries in} loader($N$)} 
      \Comment{Load batches with N samples}
      \State $y_a, y_b = augment(x)$ \Comment{ Augmentations of x}
      
      \Comment{Compute representations}
      \State $z_a = f(y_a)$   \Comment{ N*D}
      \State $z_b = f'(y_b)$ \Comment{ N*D}
      
      \Comment{Get queries and positives}
      \State $G = \left[ z_a[:,i] \text{ for i {\bfseries in} range($D$)} \right]$ 
      \State $H^+ = \left[ z_b[:,i] \text{ for i {\bfseries in} range($D$)} \right]$
      \State $L_{dimCL} = 0$
      \For{$i$ {\bfseries in} range($D$)}
     \State $H^- = G \cup H^+ \setminus \{G[i],H[i]\}$  \Comment{2D-2 elements}
     \State$L_{dimCL} = L_{dimCL}+ \mathcal{L}_{{i}}^{DimCL}$ \Comment{Equation \ref{eq:infodcl}}
      \EndFor
      \State $L_{dimCL} = L_{dimCL}/D$
      \State  $Loss = \lambda L_{dimCL} +(1-\lambda)L_{base}(z_a,z_b) $
      
      \Comment{Optimization step}
      \State  $Loss$.backward()
      \State  $Optim$.step()
      \EndFor
    \end{algorithmic}
\end{algorithm}

\section{Experiments}
\begin{table*}[!ht]
    \begin{center}
    \begin{small}
    \resizebox{1.0\hsize}{!}{
    \begin{tabular}{c|c|c|ccl|ccl}
    \toprule
    \multirow{2}{*}{Datasets} & \multirow{2}{*}{Method} & \multirow{2}{*}{Type} & \multicolumn{3}{c}{ ResNet-18 } & \multicolumn{3}{|c}{ ResNet-50 } \\ \cmidrule{4-9}
    & & & BASE & + DimCL & $\Delta_{acc}$ & BASE & + DimCL & $\Delta_{acc}$ \\
    \midrule
    \multirow{4}{*}{CIFAR-10} & MoCo v2 \cite{he2019moco}* & \multirow{2}{*}{CL} & 89.55 & \textbf{89.59} & \textcolor{Green}{+ 0.04} & 90.60 & \textbf{91.12} & \textcolor{Green}{+ 0.60} \\ 
    & SimCLR \cite{chen2020simple} & & 86.01 & \textbf{88.32} & \textcolor{Green}{+ 2.31} & 86.71 & \textbf{89.67} & \textcolor{Green}{+ 2.96 } \\ \cmidrule{2-9}
    & BYOL \cite{grill2020bootstrap} & \multirow{2}{*}{Non-CL} & 88.51 & \textbf{90.57} & \textcolor{Green}{+ 2.06} & 88.0 & \textbf{89.98} & \textcolor{Green}{+ 1.98} \\
    & SimSiam \cite{chen2021exploring} & & 83.33 & \textbf{88.22} & \textcolor{Green}{+ 4.89} & 84.60 & \textbf{89.33} & \textcolor{Green}{+ 4.73 } \\
    \midrule
    \multirow{4}{*}{CIFAR-100} & MoCo v2 \cite{he2019moco}* & \multirow{2}{*}{CL} & 62.79 & \textbf{64.04} & \textcolor{Green}{+ 1.25} & 64.68 & \textbf{66.24} & \textcolor{Green}{+ 1.56} \\
    & SimCLR \cite{chen2020simple} & & 58.21 & \textbf{61.75} & \textcolor{Green}{+ 3.54} & 60.81 & \textbf{65.27} & \textcolor{Green}{+ 4.46} \\ \cmidrule{2-9}
    & BYOL \cite{grill2020bootstrap} & \multirow{2}{*}{Non-CL} & 62.36 & \textbf{67.85} & \textcolor{Green}{+ 5.49} & 64.71 & \textbf{70.94} & \textcolor{Green}{+ 6.23} \\
    & SimSiam \cite{chen2021exploring} & & 51.67 & \textbf{62.49} & \textcolor{Green}{+ 10.82} & 54.00 & \textbf{65.40} & \textcolor{Green}{+ 11.4} \\
    \midrule
    \multirow{4}{*}{STL-10} & MoCo v2 \cite{he2019moco}* & \multirow{2}{*}{CL} & 85.96 & \textbf{86.34} & \textcolor{Green}{+ 0.38} & 88.16 & \textbf{88.40} &  \textcolor{Green}{+ 0.24 } \\
    & SimCLR \cite{chen2020simple} & & 82.35 & \textbf{82.73} & \textcolor{Green}{+ 0.48 } & 84.33 & \textbf{86.28} & \textcolor{Green}{+ 1.95} \\ \cmidrule{2-9}
    & BYOL \cite{grill2020bootstrap} & \multirow{2}{*}{Non-CL} & 83.36 & \textbf{84.94} & \textcolor{Green}{+ 1.58} & 83.83 & \textbf{87.89} & \textcolor{Green}{+ 4.05} \\
    & SimSiam \cite{chen2021exploring} & & 84.24 & \textbf{84.35} & \textcolor{Green}{+ 0.11} & 86.13 & \textbf{87.14} & \textcolor{Green}{+ 1.01} \\
    \midrule
    \multirow{4}{*}{ImageNet-100} 
    & MoCo v2 \cite{he2019moco}* & \multirow{2}{*}{CL} & 76.02 & \textbf{78.38} & \textcolor{Green}{+ 2.36} & 82.36 & \textbf{83.18} & \textcolor{Green}{+ 0.82 } \\
    & SimCLR \cite{chen2020simple} &  & 75.96 & \textbf{76.52} & \textcolor{Green}{+ 0.56} & 80.86 & \textbf{81.78} & \textcolor{Green}{+ 1.12} \\ \cmidrule{2-9}
    & BYOL \cite{grill2020bootstrap} & \multirow{2}{*}{Non-CL} & 77.30 & \textbf{80.72} & \textcolor{Green}{+ 3.42} & 81.74 & \textbf{84.80 } & \textcolor{Green}{+ 3.06 } \\
    & SimSiam \cite{chen2021exploring} &  & 70.64 & \textbf{76.08} & \textcolor{Green}{+ 5.42} & 72.98 & \textbf{80.20} & \textcolor{Green}{+ 7.22} \\
    \bottomrule
    \end{tabular} }
    \end{small}
    \end{center}
    \caption{ The top-1 classification test accuracy (\%) of the BASEs (the baseline frameworks) + DimCL (the baseline with DimCL regularization) amongst various datasets, and backbones. All models are trained for 200 epochs
    Classification is performed with a linear classifier trained on top of the frozen pre-trained encoder (output of the evaluated framework).  
    ``*'' denotes an improved version of MoCo v2 with symmetric loss. }
    \label{tab:acc_dataset_archictecture}
\end{table*}

\subsection{Experiment Setup}
To show its effectiveness, we evaluate DimCL by assimilating it to state-of-the-art non-CL and CL-based frameworks. Five widely used benchmark datasets are considered including CIFAR-10 \cite{krizhevsky2009learning}, CIFAR-100 \cite{krizhevsky2009learning}, STL-10 \cite{coates2011analysis}, ImagetNet-100 \cite{tian2019contrastive}, and ImageNet-1K (1000 classes) \cite{krizhevsky2012imagenet}. Different encoders (ResNet-18, ResNet-50) are also considered. The performance is bench-marked with linear classification evaluation and transfer learning with object detection following the common evaluation protocol in  \cite{grill2020bootstrap,he2020momentum,chen2021exploring}. To be more specific, the encoder is pre-trained in an unsupervised manner on the training set of the selected dataset without labels \cite{krizhevsky2012imagenet}. 
For the linear classification evaluation, the pre-trained frozen encoder is evaluated by training an additional linear classifier and tested on the corresponding test set. 
For object detection evaluation, the pre-trained frozen encoder is evaluated by a Faster R-CNN detector (C4-backbone) with the object detection datasets (i.e., VOC object detection). 
In this paper, the Faster R-CNN detector (C4-backbone) is finetuned on the VOC train-val 07+12 set with standard 2x schedule and tested on the VOC test2007 set \cite{wang2020DenseCL,he2020momentum}.
More details regarding the two evaluation methods are provided in Appendix

\subsection{Implementation Details}


For a simple implementation, DimCL directly uses the InfoNCE loss \cite{chen2020simple} but transposes the input. BCL framework implementations are based on the open library solo-learn \cite{turrisi2021solo}. Setups of the SSL baseline framework for training are described below.

\textbf{Image augmentations.} The paper follows the setting in previous approach \cite{grill2020bootstrap, chen2020simple}. Concretely, a patch of the image is sampled and resized to 224 × 224. Random horizontal flips and color distortion are applied in turn. The color distortion is a random sequence of saturation, contrast, brightness, hue adjustments, and an optional grayscale conversion. Gaussian blur and solarization are applied to the patches at the final. 

\textbf{Training.} We use stochastic gradient descent (SGD) as the optimizer. The SGD weight decay is set to 1e-5, and the SGD momentum is 0.9 as BYOL \cite{grill2020bootstrap}. We use a batch size of 256, and a single GPU for all methods except in benchmark ImageNet-1K for which a mini-batch size of $64 \times 8$ to train on an 8-GPUs machine (NVIDIA Titan Xp) is used. As a standard practice, the learning rate is decayed using the cosine scheduler with ten epochs warm-up at the beginning \cite{loshchilov2016sgdr}. For baselines, we use the optimal set of hyperparameters tuned by \cite{turrisi2021solo}. We re-train all baselines in the same environment for a fair comparison. 
The balance weight factor $\lambda$ is set to 0.1. The temperature $\tau$ is set to 0.1 for all experiments.

\subsection{Experimental Results}
This session reports the results with four settings to prove the efficacy of DimCL: (1) Compatibility and generalization (experiments are performed with 200 epochs across non-CL (and CL) frameworks, datasets, and backbones) (2) Large-scale dataset (experiments are performed with 100 epochs on ImageNet-1K) (3) Longer Training (experiments are conducted with 1000 epochs on CIFAR-100, ImageNet-100), and (4) Transfer Learning on Object Detection.

\subsubsection{Compatibility and generalization}
To show the compatibility with various SSL frameworks and generalization across datasets and backbones, we provide the extensive result as shown in Tab. \ref{tab:acc_dataset_archictecture}. 
The results demonstrate that assimilating DimCL consistently improves the performance by a large margin for all frameworks (MoCo v2, SimCLR, BYOL, SimSiam), datasets (CIFAR-10, CIFAR-100, STL-10, ImageNet-100), and backbones (ResNet-18, ResNet-50). 

For example, on CIFAR-100 with Resnet-50, DimCL enhances the baseline MoCo v2 and SimCLR with a performance boost of  1.56\% and +4.46\%, respectively. A more significant performance boost can be observed for BYOL (+6.63\%), and SimSiam (+11.4\%). In addition, during the experiment, the BASEs are highly tuned to get the best performance, and BASEs+DimCL does not. With a fine-tuned parameter search, a higher gain might be possible. Overall, the result indicates DimCL is compatible with both CL and non-CL SSL frameworks with a non-trivial performance gain. Furthermore, it also has good generalization across various datasets and backbones.

When evaluating the performance of DimCL under different metrics, the result suggests the same conclusion. To be more specific, an experiment is conducted on  CIFAR100 with Resnet-18 backbone.  The pre-trained models of the baselines and DimCL are evaluated on the classification task with different performance metrics: Top-1 Accuracy, Top-5 Accuracy, Top-1 KNN, and Top-5 KNN. The result, shown in Tab. \ref{tab:moremetrics}, suggests that DimCL consistently improves the baseline under various performance metrics.
\begin{table*}[!ht]
\centering
\begin{tabular}{@{}l|cc|cc|cc|cc@{}}
\toprule
Metrics & \multicolumn{2}{c|}{Top-1 Acc} & \multicolumn{2}{c|}{Top-5 Acc} & \multicolumn{2}{c|}{Top-1 KNN} & \multicolumn{2}{c}{Top-5 KNN} \\ \midrule
Methods & BASE      & +DimCL             & BASE     & +DimCL            & BASE      & +DimCL             & BASE     & +DimCL           \\ \midrule
MoCo V2 & 62.79     & \textbf{64.04}     & 88.75     & \textbf{89.3}      & 57.16     & \textbf{57.82}     & 80.81     & \textbf{81.61}    \\ \midrule
SimCLR  & 58.21     & \textbf{61.75}     & 84.97     & \textbf{87.72}     & 51.7      & \textbf{54.85}     & 76.61     & \textbf{77.82}    \\ \midrule
BYOL    & 62.36     & \textbf{67.85}     & 88.51     & \textbf{90.87}     & 56.2      & \textbf{59.91}     & 80.27     & \textbf{81.94}    \\ \midrule
SimSiam & 51.67     & \textbf{62.49}     & 81.65     & \textbf{88.17}     & 50.11     & \textbf{55.12}     & 76.19     & \textbf{79.65}    \\ \bottomrule
\end{tabular}

    \caption{Performance evaluated with different metrics. The methods are trained on the CIFAR-100 dataset with 200 epochs and use Resnet-18 as the backbone. Classification is performed with a linear classifier trained on top of the frozen pre-trained encoder. The test accuracy is reported with various performance metrics: Top-1 accuracy, Top-5 accuracy, Top-1 KNN, and Top-5 KNN.}
    \label{tab:moremetrics}
\end{table*}

\subsubsection{Large-scale dataset}

\begin{table}
    \centering
    \resizebox{0.8\hsize}{!}{
    \begin{tabular}{ccc}
    \toprule
    Method & Top-1 (\%) & Top-5 (\%) \\
    \midrule
    MoCo v2 \cite{chen2020improved}$\dag$ & 67.4 & - \\
    BYOL \cite{grill2020bootstrap}$\dag$ & 66.5 & - \\
    BYOL (Reproduce) & 67.3 & 88.0 \\
    \midrule
     BYOL + DimCL  & \textbf{69.3}  & \textbf{89.0}  \\
    \bottomrule
    \end{tabular} }
    \footnotesize
  \caption{Imagenet-1K classification. All frameworks are trained without labels on the training set for 100 epochs. Evaluation is on a single crop $224 \times 224$. ``$\dag$'' denotes the results employed from \cite{chen2021exploring}.
  }
  \label{tab:imagenet1k}
\end{table}


For the large-scale dataset,  Imagenet-1K is chosen, and BYOL is selected as the baseline. Due to the resource constraint, BYOL and BYOL+DimCL are pre-trained for 100 epochs without labels. The results are reported in Tab. \ref{tab:imagenet1k}.
The results show that on the large-scale dataset, DimCL improves the BYOL baseline with a performance boost of $+2.0\%$ and outperforms all other frameworks. The performance is consistent with the results in Tab. \ref{tab:acc_dataset_archictecture}, verifying the generalization and effectiveness of DimCL.
\subsubsection{Longer Training}
\begin{figure}
    \centering
    \subfloat[200 epochs]{\includegraphics[width=0.5\linewidth]{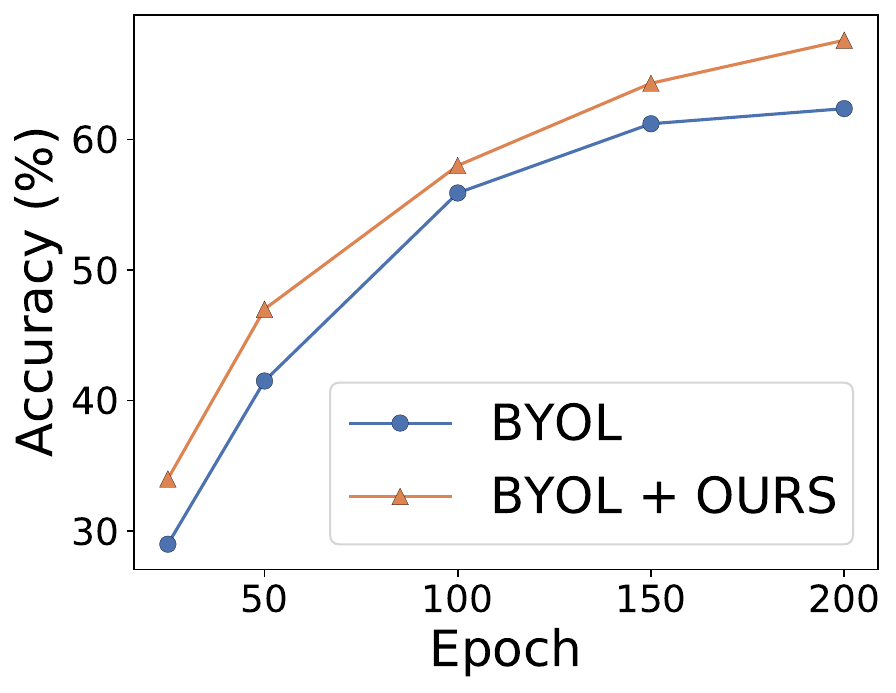}\label{fig:shorttrain}}
    \subfloat[1000 epochs]{\includegraphics[width=0.5\linewidth]{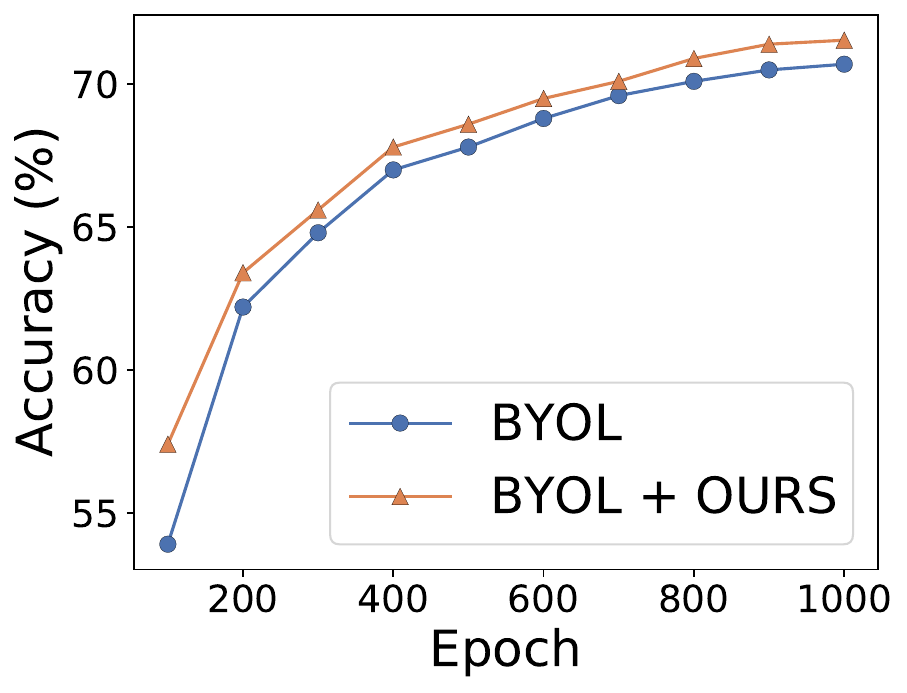}\label{fig:longtrain}}
    \caption{  Top-1 classification accuracy learning curve on the test set of CIFAR-100 of (a) 200 epochs and (b) 1000 epochs. The figure shows the consistent result between long and short training.
    Note that at the same epoch, the Top-1 accuracy of two settings is not necessarily the same due to using the cosine learning rate scheduler. }
    \label{fig:learning_curve}
\end{figure}

To demonstrate the results are consistent between short training (200 epochs) and long training (1000 epochs), we conduct experiments on CIFAR-100 and ImageNet-100 with BYOL as the baseline framework \cite{grill2020bootstrap}. Top-1 classification accuracies are reported in Tab. \ref{tab:longtraining}.

We observe that DimCL also has a consistent performance boost for the long training. Specifically, incorporating DimCL helps to significantly boost the top-1 accuracy of 
BYOL from 70.54\% to \textbf{71.94\%} (+1.4\%) for CIFAR-100, and further improves BYOL from 81.24\% to \textbf{82.51\%} for ImageNet-100.  It is reasonable that the performance boost margin can be relatively smaller in the setup of the long training compared to the short training. 

Fig. \ref{fig:learning_curve} shows the learning curve in two different settings: 200 epochs (a) and 1000 epochs (b). The results demonstrate that our method does not vanish but further improves BYOL in long training. There is a high correlation in performance improvement between short and long training. It proves that the 200 epochs setting is reasonably adequate to evaluate the performance gain.

\begin{table*}[!bpht]
\centering
\resizebox{0.7\hsize}{!}{
\begin{tabular}{cccccc}
    \toprule
    \multirow{1}{*}{Dataset} & \multirow{1}{*}{Epoch} & SimCLR & MoCo v2 & BYOL & BYOL+DimCL \\
    \midrule
    \multirow{2}{*}{CIFAR-100} & 200 & 58.21 & 62.79 & 62.36 & \textbf{67.85} \\
    & 1000 & 65.85 & 69.39 & 70.54 & \textbf{71.94} \\
    \midrule
    \multirow{2}{*}{ImageNet-100} & 200 & 75.96 & 76.02& 77.30 & \textbf{80.72} \\
    & 1000 & 78.76 & 79.98 & 81.24 & \textbf{82.51} \\
    \bottomrule
\end{tabular} }
\caption{Long training with 1000 epochs. Linear classification accuracy (\%) on the test set of CIFAR-100. All models are pre-trained on the training set without labels before evaluation. Note that MoCo v2+ is the  improved version of MoCo v2 with symmetric loss \cite{turrisi2021sololearn}.}
\label{tab:longtraining}
\end{table*}

\subsubsection{Transfer Learning on Object Detection}
\begin{table}
    \centering
    \resizebox{1\hsize}{!}{
    \begin{tabular}{ccccc}
    \toprule
   Method & Epoch & AP & AP50 & AP75 \\
    \midrule
    SimCLR \cite{chen2020simple} \dag & 200 & 51.5 & 79.4 & 55.6 \\
   BYOL \cite{grill2020bootstrap} \dag & 200 & 51.9 & 81.0 & 56.5 \\
   \midrule
   BYOL \cite{grill2020bootstrap} & 100 & 50.3 & 79.8 & 54.2 \\
   BYOL + DimCL & 100 & \textbf{55.6} & \textbf{81.9} & \textbf{61.4} \\
    \bottomrule
    \end{tabular}  }
   \caption{ Transfer learning on detection task VOC07. The \dag~ denotes the results from \cite{wang2020DenseCL}.  }
    \label{tab:detection}
\end{table}



 Tab. \ref{tab:detection} shows Objection detection evaluation on pre-trained frozen backbone ResNet-50 in Tab. \ref{tab:imagenet1k}. DimCL significantly boosts BYOL in the object detection task by a large margin. Specifically, in 100 epoch pre-training on ImageNet-1K with BYOL \cite{grill2020bootstrap}, the encoder gives 50.3, 79.8, 54.2 in AP, AP50, AP75, respectively. Doubling pre-training epochs with BYOL, \ie 200 epochs, the encoder shows a slight improvement. By contrast, in 100 epoch pre-training with BYOL+DimCL, the encoder can strongly outperform BYOL for the AP, AP50, AP75 with \textbf{55.6, 81.9, 61.4}, and even surpass the performance of the baseline 200 epoch pre-training with BYOL for all metrics. It demonstrates the effectiveness of the proposed DimCL.
Overall, the results demonstrate DimCL is effective for both downstream tasks: \textit{classification} and \textit{object detection}.



\section{Ablation study}

In this section, we provide ablation for important hyper-parameters of DimCL: the temperature $\tau$, the weight factor $\lambda$, and the dimensionality $D$.

\subsection{The effect of the temperature $\tau$}

\begin{figure}
    \centering
    \subfloat[ Feature diversity] {\includegraphics[width=0.495\linewidth]{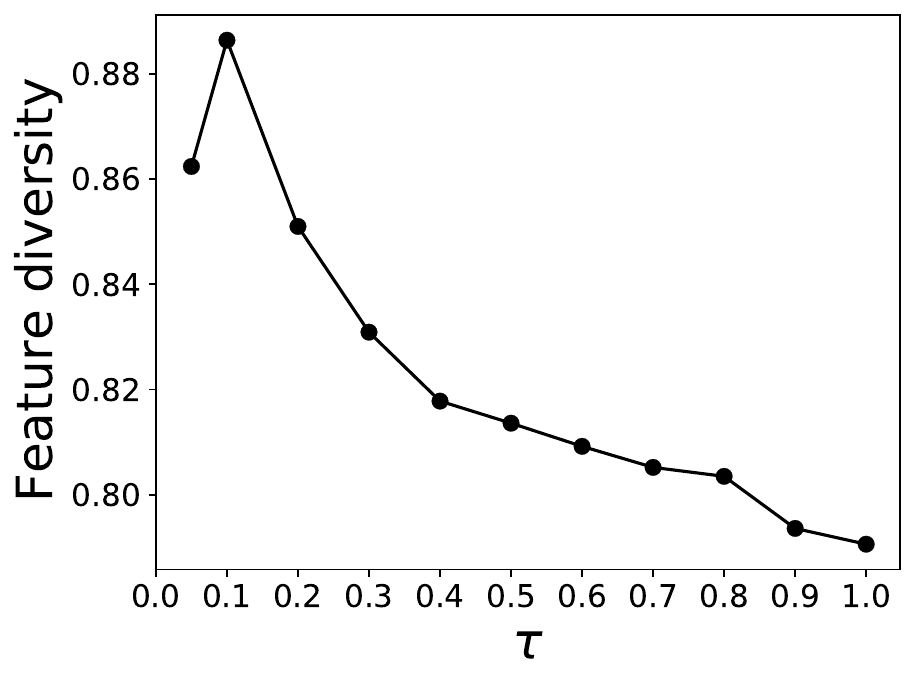}\label{fig:tau_decor_uncor}}
    \subfloat[ Top-1 linear accuracy ] {\includegraphics[width=0.48\linewidth]{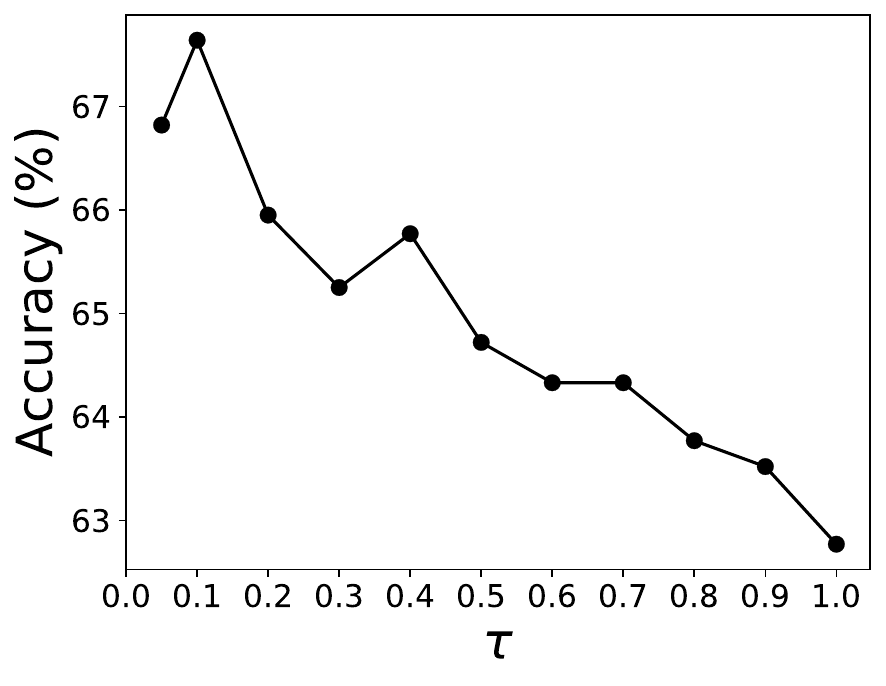}\label{fig:tau_decor_val_acc1}}
    \caption{ Feature diversity (a) and performance (b) with respect to $\tau$ on the test set of CIFAR-100. Our hypothesis emphasizes the importance of increasing the feature diversity or decreasing the correlation to remove the residual information of the feature representation. }
    \label{fig:tau_decor}
\end{figure}
We monitor the changes in feature diversity and performance when assimilating  DimCL to BYOL with various $\tau$ values. The experiment runs on CIFAR-100 for 200 epochs.
The results in Fig. \ref{fig:tau_decor} suggest that selecting a reasonable $\tau$ leads to high feature diversity (and performance). $\tau = 1$ does not lead to good feature diversity. The value of $\tau$ that supports gaining the best performance is around 0.1. This result coincides with the  $\tau$ used in conventional  BCL frameworks \cite{chen2020simple, he2020momentum}. There is a drop in performance when using too large or too small $\tau$ in DimCL.

\subsection{The effect of the weight factor $\lambda$}

\begin{figure}
    \centering
    \subfloat[Top-1 linear accuracy] {\includegraphics[width=0.49\linewidth]{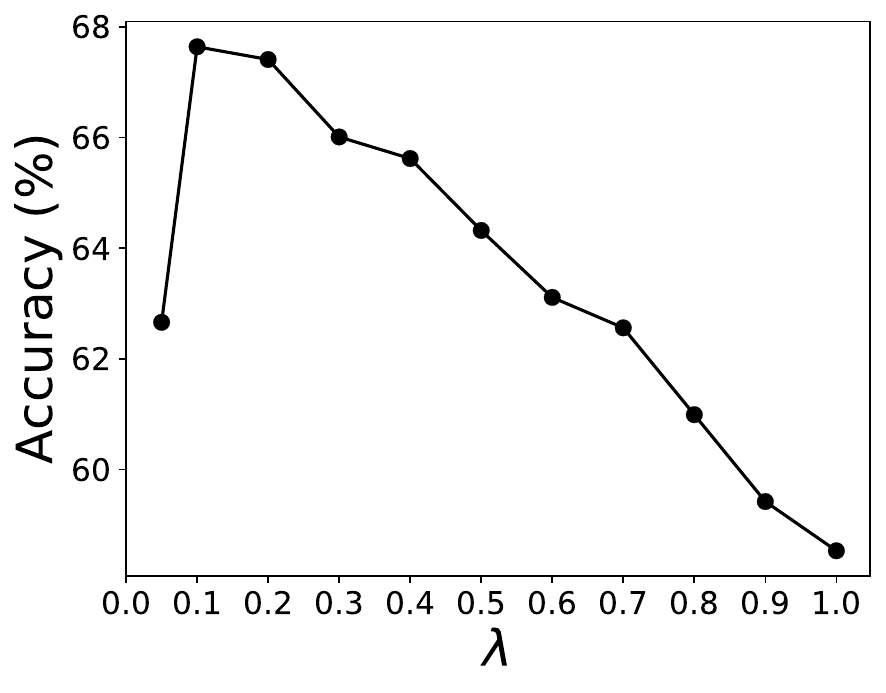}\label{fig:acc_lam_decor}}
    \subfloat[ KNN accuracy ]{\includegraphics[width=0.49\linewidth]{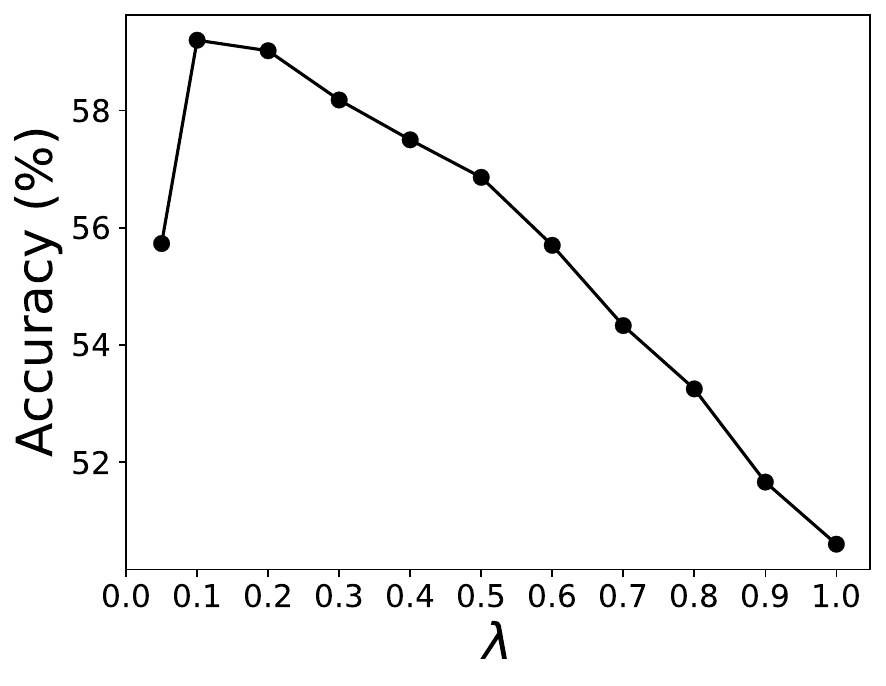}\label{fig:knn_lam_decor}}
    \caption{ Top-1 classification accuracy and top-1 KNN  with respect to $\lambda$ on the test set of CIFAR-100. Note that, the performance at $\lambda = 0$, corresponding to the performance of the baseline BYOL, is much lower than the case when incorporating BYOL with our loss. }
    \label{lbl:lam_decor}
\end{figure}

The balance weight factor between DimCL and the baseline plays an important role in gaining performance. We conduct the experiments with a range of [0, 1] for $\lambda$ in Eq. \ref{eq:finalloss}. All other parameters are kept unchanged. 

The results in Fig. \ref{lbl:lam_decor} show that with all $\lambda$ in the range of (0, 0.7), our method consistently outperforms the baseline BYOL (corresponding to $\lambda = 0$) in both two measures: top-1 classification accuracy and top-1 KNN accuracy. $\lambda = 0.1$ is found to be the optimal value to boost performance when plugging DimCL into BYOL. $\lambda$ usually depends on the baselines and dataset. However, we empirically find that setting $\lambda$ to 0.1 often gives the best performance for the most recent SSL frameworks in datasets. It is recommended to use this value at the beginning of the tuning process when using our DimCL regularization.

\subsection{The effect of the dimensionality $D$}

As the DimCL targets to address dimension-wise diversity, dimensionality should be a key fact that needs to be considered. We provide ablation studies on the effects of dimensionality. Tab. \ref{tab:dimensionality} shows the top 1 accuracy on CIFAR 100 with 200 epochs of BYOL and BYOL+DimCL. 

The result shows that for the small dimensionality, DimCL provides a large improvement over the baseline. For bigger dimensionality, the improvement tends to reduce. It is understandable since DimCL aims to maximize the useful information (or in other words, minimize the redundancy) contained in a low dimensionality. For bigger dimensionality, there is plenty of space for storing information which reduces the importance of DimCL. It is also noticeable that for very small dimensionality, the performance starts to drop for both BYOL and BYOL+DimCL (e.g: under 256) since there is not much space for storing information.

\begin{table}[!ht]
\centering
 \begin{small}
    \resizebox{1\hsize}{!}{
\begin{tabular}{@{}lccccccc@{}}
\toprule
Dimensionality & 64    & 128   & 256   & 512   & 1024  & 2048  & 8192  \\ \midrule
BYOL           & 59.85 & 60.72 & 62.36 & 62.62 & 62.44 & 62.02 & 62.99 \\
BYOL+DimCL     & 66.47 & 66.84 & 67.85 & 67.41 & 67.18 & 67.41 & 67.33 \\
Improvement    & 6.62  & 6.12  & 5.49  & 4.79  & 4.74  & 5.39  & 4.34  \\ \bottomrule
\end{tabular}}
\end{small}
\caption{The effects of DimCL on dimensionality. The table shows the top 1 accuracy on CIFAR 100 with 200 epochs of BYOL and BYOL+DimCL }
\label{tab:dimensionality}
\end{table}

\section{discussion}



\begin{table*}[!bp]
    \begin{center}
    \resizebox{0.8\hsize}{!}{
    \begin{tabular}{ccccccc}
    \toprule
    \multirow{2}{*}{Method} & \multicolumn{2}{c}{ Feature Diversity} & \multicolumn{2}{c}{ Accuracy Top-1 } & \multicolumn{2}{c}{ Accuracy KNN } \\
    & BASE & + DimCL & BASE & + DimCL & BASE & + DimCL \\
    \midrule
    BYOL \cite{grill2020bootstrap} & 0.83 & \textbf{0.88} (\textcolor{Green}{+0.05}) & 62.36 & \textbf{67.85} (\textcolor{Green}{+5.49}) & 56.20 & \textbf{59.91} (\textcolor{Green}{+3.71}) \\
    SimSiam \cite{chen2021exploring} & 0.75 & \textbf{0.92} (\textcolor{Green}{+0.17}) & 51.67 & \textbf{62.49} (\textcolor{Green}{+10.82}) & 50.11 & \textbf{55.12} (\textcolor{Green}{+5.01} ) \\
    \bottomrule
    \end{tabular} 
    }
    \end{center}
    \caption{
    Comparison of Feature diversity and performance in CIFAR-100 dataset for both BASE (baseline) and +DimCL (baseline with DimCL regularization). All frameworks are pre-trained with 200 epochs on ResNet-18 backbone.  
    }
    \label{tab:uncor_acc}
\end{table*}

\subsection{Feature Diversity Enhancement}

Our proposed DimCL is motivated to enhance feature diversity which is defined as the independence among the elements of a representation. In other words, good feature diversity means each element of representation should carry a piece of distinct information about the input image. In this view, feature diversity can be evaluated by considering correlation among all pairs of negative \textit{column vectors}. Given a tensor with size $N \times D$,  the feature diversity measure is defined as: 
\begin{equation}
    \textit{feature diversity} = 1-  \frac{1}{D(D-1)}\sum_i^D\sum_{j\neq i}^{D} |sim(g_i {\cdot}h_j)|.
\end{equation}
\begin{figure}
    \centering
    \subfloat[ Performance ] {\includegraphics[width=0.48\linewidth]{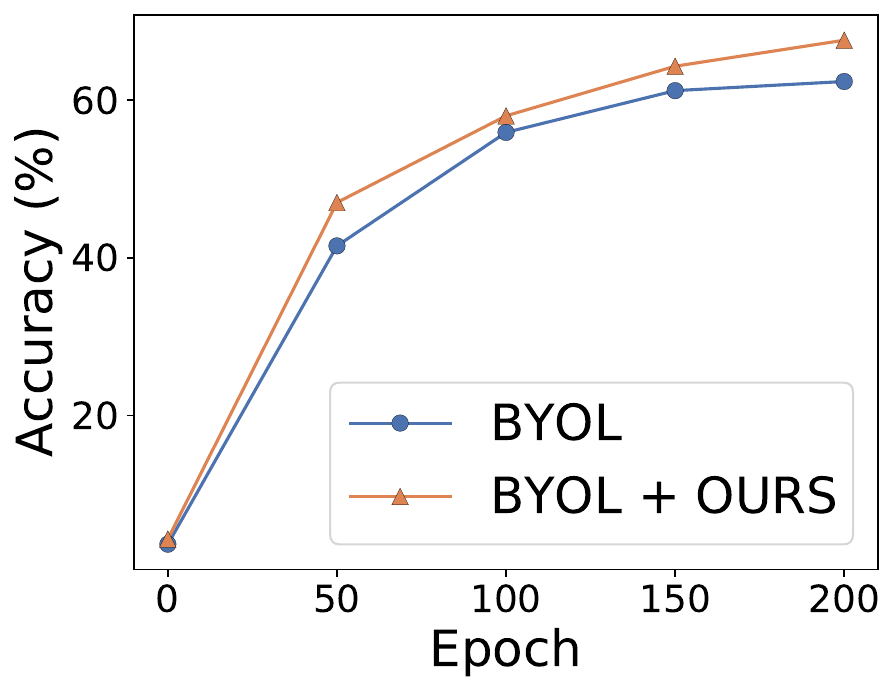}\label{fig:acc_200ep_byol_ours}}
    \subfloat[ Feature diversity] {\includegraphics[width=0.48\linewidth]{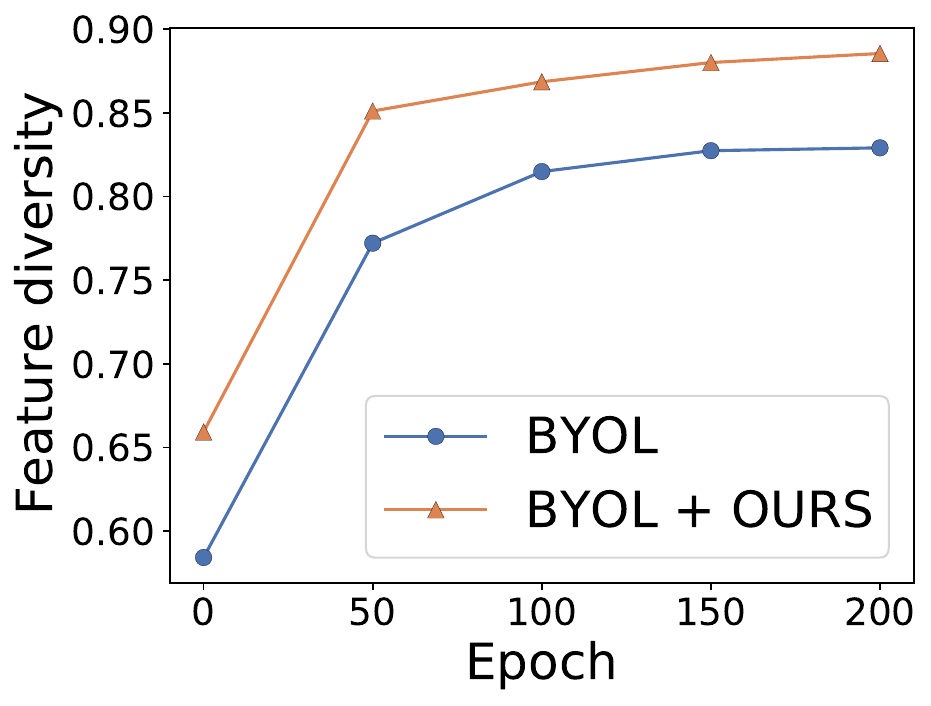}\label{fig:corr_200ep_byol_ours}}
    \caption{Relation between feature diversity and performance during training on CIFAR-100. a) the top-1 test classification accuracy. b) the corresponding feature diversity. The higher feature diversity leads to higher performance. }
    \label{fig:correlation}
\end{figure}

Here, $g, h \in \mathbb{R}^N$  are \textit{column vectors}. $sim(.)$ is the cosine similarity measure. The range for the \textit{feature diversity} measure is within $[0,1]$. The optimum value of \textit{feature diversity} is 1 which means all elements of representation are mutually independent.


To prove the enhancement of feature diversity, we take BYOL \cite{grill2020bootstrap} and SimSiam \cite{chen2021exploring} into account where the encoders are designed to learn the representation, which is invariant to augmentation without considering feature diversity. We assimilate DimCL to BYOL, and Simsiam then observe changes in feature diversity and accuracy. Results are reported in Tab. \ref{tab:uncor_acc}. 

Interestingly, the BASEs generate embedding, which already has high feature diversity.  Adding DimCL to BASEs has a strong effect on further increasing the feature diversity, which impacts performance improvement.
Specifically, DimCL makes an improvement 0.05 (5\% in percentage) feature diversity with corresponding 5.49\% accuracy on BYOL and  0.17 (17\% in percentage) feature diversity with corresponding 10.82\% accuracy on SimSiam. The more feature diversity improvement, the better performance gain. The relation between feature diversity is shown clearly in Fig. \ref{fig:correlation}. 

From the perspective of information theory, improving feature diversity can be classified as the Information Bottleneck objective \cite{tishby2015deep} which forces a representation that conserves as much information about the sample as possible. It is mentioned to be beneficial in various research \cite{ayinde2019regularizing,hua2021feature,tishby2015deep}.  Our result is one of the empirical pieces of evidence  proving the benefit of feature diversity.

\subsection{Hardness-Aware Property in DimCL}

The hardness-aware property plays a key role in BCL controlling the uniformity-tolerance dilemma \cite{wang2020understanding} leading to its success. In the view of optimization, the hardness-aware property puts more weight into optimizing negative pairs that have high similarities. This way is influenced by hard examples mining and has proven to be effective \cite{nozawa2021understanding, Zeng_2020_CVPR, Vasudeva_2021_ICCV, Wang_2021_ICCV, Bai_2021_ICCV, iscen2018mining}. 

The interpretation of Harness-aware in the DimCL can be understood via gradient analysis of loss function. Let's consider loss for query $g_i$:


\begin{equation}
\mathcal{L}_i^{DimCL} = -\log \frac{ \exp(g_i {\cdot} h^+_i /\tau ) }{\exp(g_i {\cdot} h^+_i/\tau ) + \sum_{j} \exp(g_i {\cdot} h^-_j /\tau )} 
\label{eq:hardness}
\end{equation}
where $g$ and $h$ are the $l_2$-normalized column vectors. The gradient of $\mathcal{L}_i^{DimCL}$ w.r.t query $g_i$ is derived as:

\begin{align}
    \frac{\partial \mathcal{L} _i^{DimCL}}{\partial g_i} = - &\frac{1}{\tau} (1-\frac{\exp(g_i {\cdot} h^+_i /\tau)}{\exp(g_i {\cdot} h^+_i /\tau) + \sum_{j} \exp(g_i {\cdot} h^-_j /\tau ) }).h^+_i \\ 
    +  &\frac{1}{\tau} \frac{\sum_{j} \exp(g_i {\cdot} h^-_j /\tau ) .h^-_j}{\exp(g_i {\cdot} h^+_i /\tau) + \sum_{j} \exp(g_i {\cdot} h^-_j /\tau ) } \\
    = - &\frac{1}{\tau} (1-\alpha_i'). h^+_i + \frac{1}{\tau} \sum_j \alpha_j . h^-_j, 
    \label{eq:Derivative}
 \end{align}

where $\alpha_i' = \frac{\exp(g_i {\cdot} h^+_i /\tau)}{\exp(g_i {\cdot} h^+_i /\tau) + \sum_{j} \exp(g_i {\cdot} h^-_j /\tau ) }$ can be interpreted as the probability of $g_i$ being recognized as the positive colunm vector $h^+_i$. Similarly, $\alpha_j = \frac{ \exp(g_i {\cdot} h^-_j /\tau ) }{\exp(g_i {\cdot} h^+_i /\tau) + \sum_{j} \exp(g_i {\cdot} h^-_j /\tau ) }$ can be interpreted as the probability of $g_i$ being recognized as the negative vector $h^-_j$. We can easily see that $\alpha_i'+\sum_j \alpha_j = 1$ and all $\alpha > 0$.

The Eq. \ref{eq:Derivative} reveals how DimCL makes the query similar to the positive key and dissimilar from negative keys. Concretely, if $g_i$ and $h^+_i$ are very close, the gradient of $g_i$ is very small because $1-\alpha_i' \approx 0$ and $\sum_j \alpha_j \approx 0$ (because $1-\alpha_i' \approx 0$ and $\alpha_i'+\sum_j \alpha_j = 1$)  . Thus, the optimizer does not update the query $g_i$. By contrast, if $g_i$ and $h^-_j$ are very close, the weight $\alpha_j$ is big, encouraging the optimizer to push the query far away from the corresponding negative keys.

Regarding the ability to differently treat negative keys, the gradient weight w.r.t negative keys are proportional to the exponential $\exp(\frac{g_i.h^-_j}{\tau})$. It shows that hard column pairs, where query $g_i$ is far with negative keys, are penalized more with larger $\alpha_j$. In other words, the optimizer will pay more attention to optimizing hard column pairs, which leads to better optimization results than treating them equally. This phenomenon is the hardness-awareness property of the loss \ref{eq:hardness}. The effect of the hardness-aware property in DimCL  in relation to feature diversity can be empirically seen clearly in the ablation study  Fig. \ref{fig:tau_decor}


\subsection{Beyond CL and non-CL.} 

 \begin{table}[ht]
  
    \centering
      \begin{small}
    \resizebox{1\hsize}{!}{
    \begin{tabular}{ccccc}
    \toprule
    \multirow{2}{*}{Datasets} & \multicolumn{2}{c}{ ResNet-18 } & \multicolumn{2}{c}{ ResNet-50 } \\
    & BT & + DimCL  & BT & + DimCL  \\
    \midrule
    CIFAR-10 &  88.45 & \textbf{89.21}  & 88.91 & \textbf{90.28}  \\
    CIFAR-100 &  65.61 & \textbf{66.42}  & 66.35 & \textbf{66.88}  \\
    STL-10 &  82.26 & \textbf{82.66}   & 84.99 & \textbf{85.34}  \\
    IMAGENET-100 & 78.50 & \textbf{78.72}  & 82.44 & \textbf{82.78}  \\
    \bottomrule
    \end{tabular} }
    \end{small}
    \caption{ DimCL for improving Barlow Twins (BT) \cite{zbontar2021barlow}. Frameworks are trained for 200 epochs with ResNet-18 and ResNet-50 backbone on the 4 datasets. We report top-1 linear classification (\%) accuracy. 
    }
    \label{tab:onbarlow}
\end{table}

 \begin{table}[ht]
    \centering
    \begin{tabular}{ccc}
    \toprule
    Dataset & BYOL+DimCL &  BYOL+Barlow Twins \\
    \midrule
    CIFAR-100 & \textbf{67.85} & 66.55  \\
    \bottomrule
    \end{tabular} 
    \caption{ Comparison between DimCL and Barlow Twins on top of baseline BYOL. Models are trained for 200 epochs with ResNet-18  on the CIFAR-100. We report top-1 linear classification (\%) accuracy. }
    \label{tab:comparsionBarlow}
\end{table}

Previous results show that DimCL is most beneficial in boosting the performance of CL and non-CL frameworks with a non-trivial margin.
Here, we also investigate the recent work that designed an explicit term for decorrelation, Barlow Twins (BT) \cite{zbontar2021barlow}. 

We experiment by adding the correlation-reduction loss of BT  to the previous baseline BYOL and comparing it against DimCL. The result in Tab. \ref{tab:comparsionBarlow} shows that BYOL+DimCL strongly outperforms BYOL+Barlow. Furthermore, as shown in Tab. \ref{tab:onbarlow}, When incorporate into BT, DimCL can also improve BT. 

This empirical result recommends that DimCL provides better performance than BT.

\subsection{DimCL for supervised learning.} 
Since DimCL works as a regularizer enhancing the feature diversity, it is expected to benefit other fields beyond self-supervised learning (e.g. supervised learning (SL)). This experiment utilizes DimCL to boost SL on CIFAR-100 and CIFAR-10 datasets. We use the solo-learn library \cite{turrisi2021solo} to train the supervised model with backbone ResNet-18 \cite{he2016deep}. 

 \begin{table}[ht]
    \centering
    \begin{tabular}{ccc}
    \toprule
    Dataset & Supervised &  + DimCL  (\%) \\
    \midrule
    CIFAR-100 & 70.27 & \textbf{71.68}  \\
    CIFAR-10 & 93.29 & \textbf{93.35}  \\
    \bottomrule
    \end{tabular} 
    \caption{ DimCL for improving supervised learning. Models are trained for 200 epochs with ResNet-18 and ResNet-50 backbone on the 4 datasets. We report top-1 linear classification (\%) accuracy. }
    \label{tab:supervised}
\end{table}

DimCL is assimilated with cross-entropy loss for training the model simultaneously.
Tab. \ref{tab:supervised} shows the top-1 classification accuracy on the test set. For CIFAR-10 DimCL shows slight improvement, while CIFAR-100 shows the DimCL supports to boost the conventional supervised learning from 70.27\% to \textbf{71.68\%} ($+$1.4\%), demonstrating the benefit of DimCL for SL.


\subsection{DimCL versus AbsCL}

In order to maximize the feature diversity, the considered query $g_i$ should be orthogonal with all negative keys  $\mathbb{H}^-_i$. The corresponding objective is: 

\begin{equation}
    \begin{aligned}
    \mathcal{L}^{AbsCL} &= \frac{1}{D} \sum_{i=1}^D \mathcal{L}_i^{AbsCL} 
    \quad \\
    \mathcal{L}_{{i}}^{AbsCL} &= -\log \frac{\exp(g_i{\cdot} h_i^+ /\tau  )}{\exp(g_i{\cdot}h_i^+/\tau) + \sum_{j=1}^{2D-2} \exp(|g_i {\cdot}h_j^-| /\tau)}.    
    \end{aligned}
\end{equation}

\begin{table}[!htbp]
    \centering
    \smallskip
    \resizebox{1\hsize}{!}{
    \begin{tabular}{c|cccc|cccc}
    \toprule
    \multirow{2}{*}{Datasets} & \multicolumn{4}{|c|}{ BYOL } & \multicolumn{4}{|c}{ SimSiam }  \\
    \cmidrule{2-9}
    & Baseline & $\tau$ & DimCL & AbsCL & Baseline & $\tau$ & DimCL & AbsCL \\
    \midrule
    \multirow{2}{*}{CIFAR-10} & \multirow{2}{*}{88.51} & 1 & 88.92 & 89.81 & \multirow{2}{*}{83.33} & 1 & 86.07 & 86.59 \\
    \cmidrule{3-5} \cmidrule{7-9}
    & & 0.1 & \textbf{90.57} & \textbf{90.83} & & 0.1 & \textbf{88.22} & \textbf{87.67} \\
    \midrule
    \multirow{2}{*}{CIFAR-100} & \multirow{2}{*}{62.34} & 1 & 62.77 & 66.02 & \multirow{2}{*}{51.67} & 1 & 54.67 & 57.56 \\
    \cmidrule{3-5} \cmidrule{7-9}
    & & 0.1 & \textbf{67.85} & \textbf{67.87} & & 0.1 & \textbf{62.49} & \textbf{62.96} \\
    \bottomrule
    \end{tabular}}
    \smallskip \smallskip
    \caption{DimCL versus AbsCL. We report top-1 linear test accuracy (\%) on CIFAR-10 and CIFAR-100. All methods are trained for 200 epochs. For $\tau=0.1$, all methods DimCL and AbsCL perform best and performance is almost similar.}
    \label{tab:abs_DimCL}
\end{table}

Empirically, Tab. \ref{tab:abs_DimCL} shows that the original InfoNCE is sufficient to achieve the objective without any modification (e.g., adding the absolute). It is important to note that without $\tau$, DimCL and AbsCL can outperform the baseline. However, to achieve the best performance, $\tau$ is needed to present. At the optimal $\tau=0.1$, the performance of DimCL is nearly the same as AbsCL. This phenomenon can be explained by considering the $\exp$ term and the effect of temperature  $\tau$. With small $\tau$, the $\exp(x/\tau)$ has a high weight on pushing positive value x toward zero with a corresponding high gradient but has almost no consideration on negative value x with the same magnitude due to its much smaller gradient.

\subsection{Visualization of Representation. }

\begin{figure}[!t]
	\vskip -0.1in
	\begin{center}
		\centerline{\includegraphics[width=0.95\columnwidth]{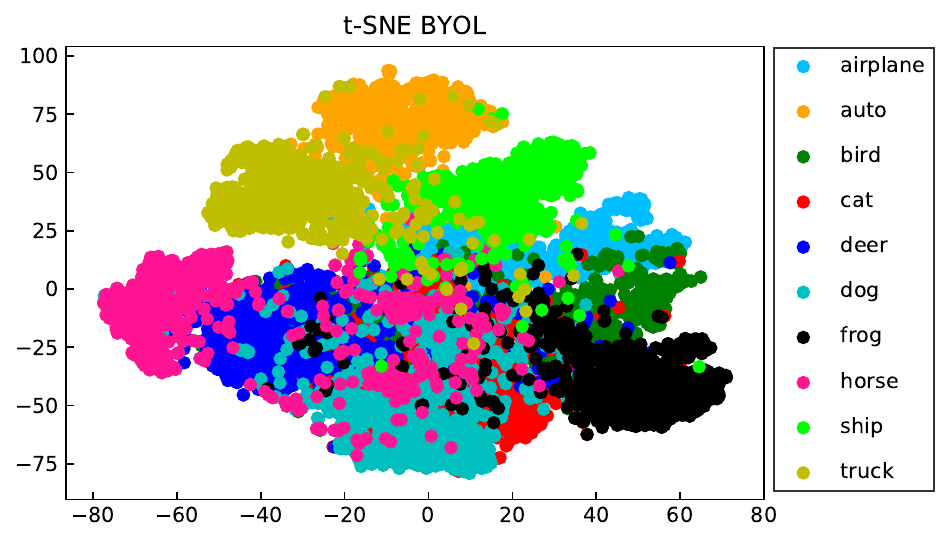}}
		\caption{ t-SNE plot of ten classes for data trained by the BYOL baseline in 200 epochs with accuracy = 88.51\% in CIFAR-10 with 10,000 samples of the test set.}
		\label{fig:tsne_byol}
	\end{center}
	\vskip -0.1in
\end{figure}

\begin{figure}[!t]
    \vskip -0.15in
	\begin{center}
		\centerline{\includegraphics[width=0.95\columnwidth]{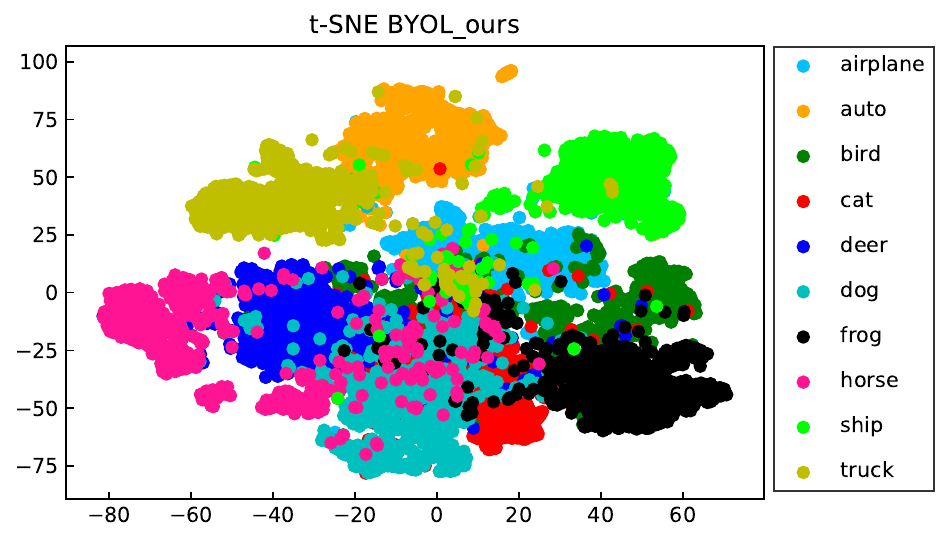}}
		\caption{  t-SNE plot of ten classes for data trained by the BYOL + DimCL in 200 epochs with accuracy = 90.57\% in CIFAR-10 with 10,000 samples of the test set.}
		\label{fig:tsne_byol_ours}
	\end{center}
	\vskip -0.1in
\end{figure}

Visualization of representation via t-SNE is reported to see the effect of DCL on representation space.  Fig. \ref{fig:tsne_byol} and Fig. \ref{fig:tsne_byol_ours} show the representation of BYOL baseline and our method on the 2D space. The experiment is conducted on CIFAR-10 with 10 classes. 
The results clearly show that our method in Fig. \ref{fig:tsne_byol_ours} gives more separable representations. More specifically, airplane, auto, ship, and truck are almost separable among them and also from other animal classes. All classes are scatted in the more compact clusters compared to the baseline in Fig. \ref{fig:tsne_byol}.

\begin{table}[htb]
\centering
    \begin{tabular}{ccc}
    \toprule
     & Intra-class distance $\downarrow$ & Inter-class distance $\uparrow$  \\
    \midrule
    BYOL & 27.0 & 82.3  \\
    BYOL+DCL & \textbf{24.1} & \textbf{85.9}  \\
    \bottomrule
    \end{tabular}  
    \caption{Inter-class distance and Inter-class distance on CIFAR-10 test set.}
    \label{tab.inter-intra}
\end{table}

To show the difference between the two representation spaces quantitatively, intra-class distance and inter-class distance \cite{taufik2014comparative} are calculated and provided in Tab. \ref{tab.inter-intra}. The quantitative result agrees that BYOL+DCL forms the more compact clusters while maintaining a higher separation among different clusters compared to BYOL.

\section{Conclusion}

This paper introduces Dimensional Contrastive Learning (DimCL), a new way of applying CL. DimCL works as a regularization that can assimilate with non-CL (and CL) based frameworks to boost performance on downstream tasks such as classification and object detection. DimCL enhances feature diversity among elements within a representation. 
DimCL has high compatibility and generalization across datasets, frameworks, and backbone architectures. We believe that feature diversity is a key indispensable ingredient for learning representation. This paper focuses on images and provides mostly empirical evidence but DimCL can be generalized to other modalities (e.g. audio, video, text) and proven with theoretical results. We let it for future work.

\bibliographystyle{plain}
\bibliography{refs.bib}

\EOD

\end{document}